%% file: main.tex
\documentclass[10pt,twocolumn,letterpaper]{article}

\usepackage[pagenumbers]{cvpr} 

\input{preamble}

\definecolor{cvprblue}{rgb}{0.21,0.49,0.74}
\usepackage[pagebackref,breaklinks,colorlinks,citecolor=cvprblue]{hyperref}

\usepackage{graphicx}
\usepackage{amsmath}
\usepackage{amssymb}
\usepackage{booktabs}
\usepackage{cuted}
\usepackage{caption}
\usepackage{cuted}
\usepackage{float}
\usepackage{graphicx}
\usepackage{pifont}
\usepackage{xcolor}
\usepackage{listings}
\usepackage{bm}

\usepackage{soul}

\def\paperID{*****} 
\def\confName{CVPR}
\def\confYear{2024}

\title{
Wonder3D: Single Image to 3D using Cross-Domain Diffusion}

\author{Xiaoxiao Long$^{1,3,6\ast}$,
Yuan-Chen Guo$^{2,3\ast}$,
Cheng Lin$^{1\dag}$,
Yuan Liu$^{1}$,
Zhiyang Dou$^{1}$
\\
Lingjie Liu$^{4}$,
Yuexin Ma$^{5}$,
Song-Hai Zhang$^{2}$,
Marc Habermann$^{6}$,
Christian Theobalt$^{6}$,
Wenping Wang$^{7\dag}$
\vspace{0.2cm}
\\
{\normalsize $^{1}$ The University of Hong Kong \quad $^{2}$ Tsinghua University}
{\normalsize \quad $^{3}$ VAST \quad }
\\
{\normalsize $^{4}$ University of Pennsylvania 
\quad $^{5}$ Shanghai Tech University
\quad $^{6}$ MPI Informatik
\quad $^{7}$
Texas A\&M University}    \\
{\normalsize {$^{\ast}$ Equal Contributions.}\quad {\href{https://www.xxlong.site/Wonder3D/}{https://www.xxlong.site/Wonder3D/}} }
}

\begin{document}
\maketitle

\let\thefootnote\relax\footnotetext{$^{\dag}$ Corresponding authors.}

\begin{strip}
    \centering
    \vspace{-5em}
    \centering
    \includegraphics[width=\textwidth]{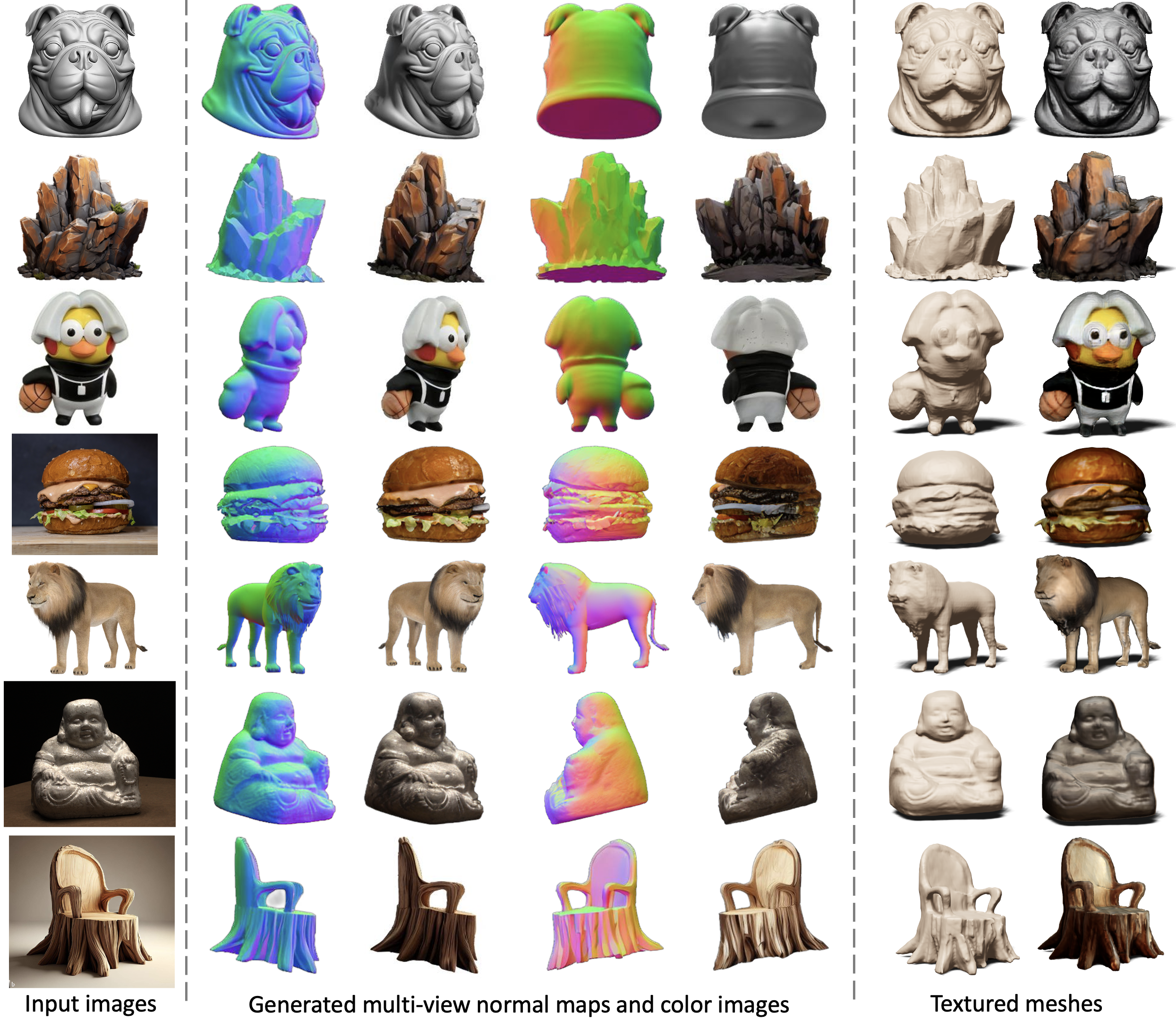}
    \vspace{-2em}
    \captionof{figure}{\textit{Wonder3D} reconstructs highly-detailed textured meshes from a single-view image in only $2 \sim 3$ minutes. 
    \textit{Wonder3D} first generates consistent multi-view normal maps with corresponding color images via a cross-domain diffusion model, and then leverages a novel normal fusion method to achieve fast and high-quality reconstruction.
    }
    \label{fig:teaser}
\end{strip}

\input{sections/0_abs}    
\input{sections/1_intro}

\input{sections/2_related_works}

\input{sections/3_method}

\input{sections/5_exp}

\input{sections/6_conclusion}

{
    \small
    \bibliographystyle{ieeenat_fullname}
    \bibliography{main}
}


\end{document}

%% file: preamble.tex
\usepackage[dvipsnames]{xcolor}


%% file: sections/0_abs.tex
\begin{abstract}

In this work, we introduce \textbf{Wonder3D}, a novel method for efficiently generating high-fidelity textured meshes from single-view images.
Recent methods based on Score Distillation Sampling (SDS) have shown the potential to recover 3D geometry from 2D diffusion priors, but they typically suffer from time-consuming per-shape optimization and inconsistent geometry. 
In contrast, certain works directly produce 3D information via fast network inferences, but their results are often of low quality and lack geometric details.
To holistically improve the quality, consistency, and efficiency of single-view reconstruction tasks, we propose a cross-domain diffusion model that generates multi-view normal maps and the corresponding color images. To ensure the consistency of generation, we employ a multi-view cross-domain attention mechanism that facilitates information exchange across views and modalities. Lastly, we introduce a geometry-aware normal fusion algorithm that extracts high-quality surfaces from the multi-view 2D representations.
Our extensive evaluations demonstrate that our method achieves high-quality reconstruction results, robust generalization, and good efficiency compared to prior works.
 
\end{abstract}

%% file: sections/1_intro.tex
\section{Introduction}
Reconstructing 3D geometry from a single image stands as a fundamental task in computer graphics and 3D computer vision~\cite{liu2023syncdreamer, jun2023shap, qian2023magic123, melas2023realfusion, dou2023tore, liu2023zero,Long_2021_ICCV, nichol2022point}, 
offering a wide range of versatile applications such as virtual reality, video games, 3D content creation, and the precision of robotics grasping.
However, this task is notably challenging since it is ill-posed and demands the ability to discern the 3D geometry of both visible and invisible parts. This ability requires extensive knowledge of the 3D world.

Recently, the field of 3D generation has experienced rapid and flourishing development with the introduction of diffusion models. A growing body of research~\cite{poole2022dreamfusion,wang2023score,wang2023prolificdreamer,lin2023magic3d,chen2023fantasia3d}, such as DreamField~\cite{jain2022zero}, DreamFusion~\cite{poole2022dreamfusion}, and Magic3D~\cite{lin2023magic3d}, resort to distilling prior knowledge of 2D image diffusion models or vision language models to create 3D models from text or images via Score Distillation Sampling (SDS)~\cite{poole2022dreamfusion}.
Despite their compelling results, these methods suffer from two main limitations: 
\textbf{efficiency} and \textbf{consistency}. The per-shape optimization process typically entails tens of thousands of iterations, involving full-image volume rendering and inferences of the diffusion models. Consequently, it often consumes tens of minutes or even hours on per-shape optimization. Moreover, the 2D prior model operates by considering only a single view at each iteration and strives to make every view resemble the input image. This often results in the generation of 3D shapes exhibiting inconsistencies, thus, often leading to the generation of 3D shapes with inconsistencies such as multiple faces (i.e., the Janus problem~\cite{poole2022dreamfusion}).

There exists another group of works that endeavor to directly produce 3D geometries like point clouds~\cite{nichol2022point, zeng2022lion, luo2021diffusion, zhou20213d}, meshes~\cite{liu2023meshdiffusion,gao2022get3d}, neural fields~\cite{wang2023rodin,cheng2023sdfusion,karnewar2023holofusion, kim2023neuralfield, gu2023learning, anciukevivcius2023renderdiffusion, muller2023diffrf, ntavelis2023autodecoding, jun2023shap, zhang20233dshape2vecset, gupta20233dgen, erkocc2023hyperdiffusion, chen2023single} via network inference to avoid time-consuming per-shape optimization. Most of them attempt to train 3D generative diffusion models from scratch on 3D assets. However, due to the limited size of publicly available 3D datasets, these methods demonstrate poor \textbf{generalizability}, most of which can only generate shapes on specific categories.

More recently, several methods have emerged that directly generate multi-view 2D images, with representative works including SyncDreamer~\cite{liu2023syncdreamer} and MVDream~\cite{shi2023mvdream}. By enhancing the multi-view consistency of image generation, these methods can recover 3D shapes from the generated multi-view images. Following these works, our method also adopts a multi-view generation scheme to favor the flexibility and efficiency of 2D representations. However, due to only relying on color images, the \textbf{fidelity} of the generated shapes is not well-maintained, and they struggle to recover geometric details or come with enormous computational costs.

To better address the issues of fidelity, consistency, generalizability and efficiency in the aforementioned works, in this paper, we introduce a new approach to the task of single-view 3D reconstruction by generating multi-view consistent normal maps and their corresponding color images with a cross-domain diffusion model.
The key idea is to extend the stable diffusion framework to model the joint distribution of two different domains, i.e., normals and colors. We demonstrate that this can be achieved by introducing a domain switcher and a cross-domain attention scheme. In particular, the domain switcher allows the diffusion model to generate either normal maps or color images, while the cross-domain attention mechanisms assist in the information exchange between the two domains, ultimately improving consistency and quality. Finally, in order to stably extract surfaces from the generated views, we propose a geometry-aware normal fusion algorithm that is robust to the inaccuracies and capable of reconstructing clean and high-quality geometries (see Figure~\ref{fig:teaser}).

We conduct extensive experiments on the Google Scanned Object dataset~\cite{downs2022google} and various 2D images with different styles.
The experiments validate that \textit{Wonder3D} is capable of producing high-quality geometry with high efficiency in comparison with baseline methods. \textit{Wonder3D} possesses several distinctive properties and accordingly has the following contributions:
\begin{itemize}

\item Wonder3D holistically considers the issues of generation quality, efficiency, generalizability, and consistency for single-view 3D reconstruction. It has achieved a leading level of geometric details with reasonably good efficiency among current zero-shot single-view reconstruction methods.

\item We propose a new multi-view cross-domain 2D diffusion model to predict normal maps and color images. This representation not only adapts to the original data distribution of Stable Diffusion model but also effectively captures the rich surface details of the target shape.

\item We propose a cross-domain attention mechanism to produce multi-view normal maps and color images that are consistently aligned. This mechanism facilitates information perception across different domains, enabling our method to recover high-fidelity geometry.

\item We introduce a novel geometry-aware normal fusion algorithm that can robustly extract surfaces from the generated normal maps and color images.

\end{itemize}

%% file: sections/2_related_works.tex
\input{figures/pipeline}

\section{Related Works}
\subsection{2D Diffusion Models for 3D Generation}
Recent compelling successes in 2D diffusion models~\cite{ho2020denoising,rombach2022high,croitoru2023diffusion} and large vision language models (e.g., CLIP model~\cite{radford2021learning}) provide new possibilities for generating 3D assets using the strong priors of 2D diffusion models.
Pioneering works DreamFusion~\cite{poole2022dreamfusion} and SJC~\cite{wang2023score} propose to distill a 2D text-to-image generation model to generate 3D shapes from texts, and many follow-up works follow such per-shape optimization scheme. 
For the task of text-to-3D~\cite{chen2023fantasia3d,wang2023prolificdreamer,seo2023ditto,yu2023points,lin2023magic3d,seo2023let,tsalicoglou2023textmesh,zhu2023hifa,huang2023dreamtime,armandpour2023re,wu2023hd,chen2023it3d} or image-to-3D synthesis~\cite{tang2023make,melas2023realfusion,qian2023magic123,xu2022neurallift,raj2023dreambooth3d,shen2023anything}, these methods typically optimize a 3D representation (i.e., NeRF, mesh, or SDF), and then leverage neural rendering to generate 2D images from various viewpoints. The images are then fed into the 2D diffusion models or CLIP model for calculating SDS~\cite{poole2022dreamfusion} losses, which can guide the 3D shape optimization.

However, most of these methods always suffer from low efficiency and multi-face problem, where a per-shape optimization consumes tens of minutes and the optimized geometry tends to produce multiple faces due to the lack of explicit 3D supervision.
A recent work one-2-3-45~\cite{hugging2023one2345} proposes to leverage a generalizable neural reconstruction method SparseNeuS~\cite{long2022sparseneus} to directly produce 3D geometry from the generated images from zero123~\cite{liu2023zero}. Although the method achieves high efficiency, its results are of low-quality and lack geometric details.

\subsection{3D Generative Models}
Instead of performing a time-consuming per-shape optimization guided by 2D diffusion models, some works attempt to directly train 3D diffusion models based on various 3D representations, like point clouds~\cite{nichol2022point, zeng2022lion, luo2021diffusion, zhou20213d}, meshes~\cite{liu2023meshdiffusion,gao2022get3d}, neural fields~\cite{wang2023rodin,cheng2023sdfusion,karnewar2023holofusion, kim2023neuralfield, gu2023learning, anciukevivcius2023renderdiffusion, muller2023diffrf, ntavelis2023autodecoding, jun2023shap, zhang20233dshape2vecset, gupta20233dgen, erkocc2023hyperdiffusion, chen2023single}
However, due to the limited size of public available 3D assets dataset, most of the works have only been validated on limited categories of shapes, and how to scale up on large datasets is still an open problem.
On the contrary, our method adopts 2D representations and, thus, can be built upon the 2D diffusion models~\cite{rombach2022high} whose pre-trained priors significantly facilitate zero-shot generalization ability.

\subsection{Multi-view Diffusion Models}
To generate consistent multi-view images, some efforts~\cite{watson2022novel,gu2023nerfdiff,deng2023nerdi,zhou2023sparsefusion,tseng2023consistent,chan2023generative,yu2023long,tewari2023diffusion,yoo2023dreamsparse,szymanowicz2023viewset,tang2023mvdiffusion,xiang20233d,liu2023deceptive,lei2022generative} are made to extend 2D diffusion models from single-view images to multi-view images.
However, most of these methods focus on image generation and are not designed for 3D reconstruction.
The works~\cite{xiang20233d,zhang2023text2nerf} first warp estimated depth maps to produce incomplete novel view images to then perform inpainting on them, but their result quality significantly degrades when the depth maps estimated by external depth estimation models are inaccurate.
The recent works Viewset Diffusion~\cite{szymanowicz2023viewset}, SyncDreamer~\cite{liu2023syncdreamer}, and MVDream~\cite{shi2023mvdream} share a similar idea to produce consistent multi-view color images via attention layers. However, 
unlike that normal maps explicitly encode geometric information, reconstruction from color images always suffers from texture ambiguity, and, thus, they either struggle to recover geometric details or require huge computational costs.
SyncDreamer~\cite{liu2023syncdreamer} requires dense views for 3D reconstruction, but still suffers from low-quality geometry and blurring textures.
MVDream~\cite{shi2023mvdream} still resorts to a time-consuming optimization using SDS loss for 3D reconstruction, and its multi-view distillation scheme requires 1.5 hours.
In contrast, our method can reconstruct high-quality textured meshes in just 2 minutes.

%% file: figures/pipeline.tex
\begin{figure*}[tp!]
\centering
\includegraphics[width=\linewidth]{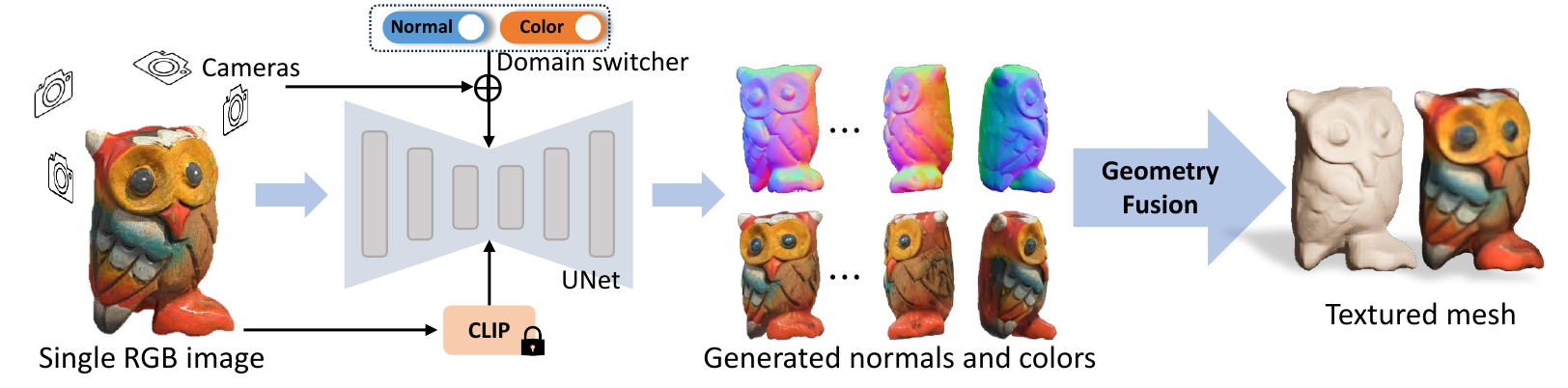}
\caption{Overview of \textit{Wonder3D}. Given a single image, \textit{Wonder3D} takes the input image, the text embedding produced by CLIP model~\cite{radford2021learning}, the camera parameters of multiple views, and a domain switcher as conditioning to generate consistent multi-view normal maps and color images. Subsequently, \textit{Wonder3D} employs an innovative normal fusion algorithm to robustly reconstruct high-quality 3D geometry from the 2D representations, yielding high-fidelity textured meshes. 
}
\label{fig:pipeline}	
\end{figure*}

%% file: sections/3_method.tex
\section{Problem Formulation}

\subsection{Diffusion Models}
Diffusion models~\cite{ho2020denoising,sohl2015deep} are first proposed to gradually recover images from a specifically designed degradation process, where a forward Markov chain and a Reverse Markov chain are adopted.
Given a sample $z_0$ drawn from the data distribution $p(z)$, the forward process of denoising diffusion
models yields a sequence of noised data $\left\{z_t \mid t \in(0, T)\right\}$ with $z_t=\alpha_t z_0+\sigma_t \epsilon$, where $\epsilon$ is random noise drawn from distribution $\mathcal{N}(0,1)$, and $\alpha_t, \sigma_t$ are fixed sequence of the noise schedule. The forward process will be iteratively applied to the target image until the image becomes complete Gaussian noise at the end. On the contrary, the reverse chain then is employed to iteratively denoise the corrupted image, i.e., recovering $z_{t-1}$ from $z_t$ by predicting the added random noise $\epsilon$.
The readers can refer to ~\cite{ho2020denoising,sohl2015deep} for more details about image diffusion models.

\subsection{The Distribution of 3D Assets}
\label{3d_distribution}
Unlike that prior works adopt 3D representations like point clouds, tri-planes, or neural radiance fields, we propose that the distribution of 3D assets, denoted
as $p_a(\mathbf{z})$, can be modeled as a joint distribution of its
corresponding 2D multi-view normal maps and corresponding color images.  Specifically, given a set of cameras $\left\{\boldsymbol{\pi}_1, \boldsymbol{\pi}_2, \cdots, \boldsymbol{\pi}_K\right\}$ and a conditional input image $y$, we have
\begin{equation}
\label{eqn:joint_distribution}
p_a(\mathbf{z})=p_{nc}\left(n^{1:K}, x^{1:K} | y \right),
\end{equation}
where $p_{nc}$ is the distribution of the normal maps $n^{1:K}$ and color images $x^{1:K}$ observed from 3D
assets conditioned on an image $y$. 
For simplicity, we omit the symbol $y$ for this equation in the following discussions.
Therefore, our goal is to learn a model $f$ that synthesizes multiple normal maps and color images of a set of camera poses denoted as
\begin{equation}
\label{eqn: model_func_0}
(n^{1:K}, x^{1:K})=f(y, \boldsymbol{\pi}_{1:K}).
\end{equation}

Adopting the 2D representation enables our method to be built upon the 2D diffusion models trained on billions of images like the Stable Diffusion model~\cite{radford2021learning}, where strong priors facilitate zero-shot generalization ability. 
On the other hand, the normal map characterizes the undulations and variations present on the surface of the shape, thus encoding rich detailed geometric information. This allows for the high-fidelity extraction of 3D geometry from 2D normal maps.


Finally, we can formulate this cross-domain joint distribution as a Markov chain within the diffusion scheme:
\begin{equation}
    p\left({n}^{(1:K)}_T, {x}^{(1:K)}_T\right) \prod_t p_\theta\left({n}^{(1:K)}_{t-1}, {x}^{(1:K)}_{t-1}|{n}^{(1:K)}_{t}, {x}^{(1:K)}_{t}\right),
\end{equation}
where $p\left({n}^{(1:K)}_T, {x}^{(1:K)}_T\right)$ are Gaussian noises. Our key problem is to characterize the distribution $p_\theta$, so that we can sample from this Markov chain to generate normal maps and images.

\section{Method}
As per our problem formulation in Section~\ref{3d_distribution}, we propose a multi-view cross-domain diffusion scheme, which operates on two distinct domains to generate multi-view consistent normal maps and color images. The overview of our method is presented in Figure~\ref{fig:pipeline}.
First, our method adopts a multi-view diffusion scheme to generate multi-view normal maps and color images, and enforces the consistency across different views using multi-view attentions (see Section~\ref{sec:mvattn}).
Second, our proposed domain switcher allows the diffusion model to operate on more than one domain while its formulation does not require a re-training of an existing (potentially single domain) diffusion model such as Stable Diffusion~\cite{radford2021learning}.
Thus, we can leverage the generalizability of large foundational models, which are trained on a large corpus of data.
A cross-domain attention is proposed to propagate information between the normal domain and color image domain ensuring geometric and visual coherence between the two domains (see Section~\ref{sec:cross-domain}).
Finally, our novel geometry-aware normal fusion reconstructs the high-quality geometry and appearance from the multi-view 2D normal and color images (see Section~\ref{sec:mesh-extraction}).

\input{figures/color-compare}

\subsection{Consistent Multi-view Generation}
\label{sec:mvattn}
The prior 2D diffusion models~\cite{radford2021learning,liu2023zero} generate each image separately, so that the resulting images are not geometrically and visually consistent across different views. 
To enhance consistency among different views, similar to prior works such as SyncDreamer~\cite{liu2023syncdreamer} and MVDream~\cite{shi2023mvdream}, we utilize attention mechanism to facilitate information propagation across different views, implicitly encoding multi-view dependencies (as illustrated in Figure~\ref{fig:trans_block}).

This is achieved by extending the original self-attention layers to be global-aware, allowing connections to other views within the attention layers. Keys and values from different views are connected to each other to facilitate the exchange of information. By sharing information across different views within the attention layers, the diffusion model perceives multi-view correlation and becomes capable of generating consistent multi-view color images and normal maps.

\subsection{Cross-Domain Diffusion}
\label{sec:cross-domain}
Our model is built upon pre-trained 2D stable diffusion models~\cite{radford2021learning} to leverage its strong generalization. 
However, current 2D diffusion models~\cite{radford2021learning,liu2023zero} are designed for a single domain, so the main challenge is how to effectively extend stable diffusion models that are capable of operating on more than one domain.

\noindent
\textbf{Naive Solutions.} To achieve this goal, we explore several possible designs. A straightforward solution is to add four more channels to the output of the UNet module representing the extra domain. Therefore, the diffusion model can simultaneously output normals and color image domains. However, we notice that such a design suffers from low convergence speed and poor generalization. This is because the channel expansion may perturb the pre-trained weights of stable diffusion models and therefore cause catastrophic model forgetting.

Revisiting Eq.~\ref{eqn:joint_distribution}, it is possible to factor the joint distribution into two conditional distributions:
\begin{equation}
\begin{aligned}
q_a(\mathbf{z})= & q_n (n^{1:K}) \cdot \\
& q_c\left(x^{1:K} \mid n^{1:K}\right) .
\end{aligned}
\end{equation}

This equation suggests an alternative solution where we could initially train a diffusion model to generate normal maps and then train another diffusion model to produce color images, conditioning on the generated normal maps (or vice versa). 
Nonetheless, the implementation of this two-stage framework introduces certain complications. It not only substantially increases the computational cost but also results in performance degradation. Please refer to Section \ref{sec:discussion} for an in-depth discussion.

\noindent
\textbf{Domain Switcher.} To overcome these difficulties mentioned above, we design a cross-domain diffusion scheme via a \textbf{\textit{domain switcher}}, denoted as $s$.
The switcher $s$ is a one-dimensional vector that labels different domains, and we further feed the switcher into the diffusion model as an extra input.
Therefore, the formulation of Eq.~\ref{eqn: model_func_0} can be extended as:
\begin{equation}
\begin{aligned}
    n^{1:K}, x^{1:K}=f(y, \boldsymbol{\pi}_{1:K}, s_{n}), f(y, \boldsymbol{\pi}_{1:K}, s_{c}).
\end{aligned}
\label{eqn: model_func_1}
\end{equation}
The domain switcher $s$ is first encoded via positional encoding~\cite{mildenhall2020nerf} and subsequently concatenated with the time embedding. This combined representation is then injected into the UNet of the stable diffusion models. Interestingly, experiments show that this subtle modification does not significantly alter the pre-trained priors. As a result, it allows for fast convergence and robust generalization, without requiring substantial changes to the stable diffusion models.

\input{figures/trans-block}
\input{figures/cartoon}

\noindent
\textbf{Cross-domain Attention.} 
Using the proposed domain switcher, the diffusion model can generate two different domains. However, it is important to note that for a single view, there is no guarantee that the generated color image and the normal map will be geometrically consistent.
To address this issue and ensure the consistency between the generated normal maps and color images, we introduce a cross-domain attention mechanism to facilitate the exchange of information between the two domains. This mechanism aims to ensure that the generated outputs align well in terms of geometry and appearance.

The cross-domain attention layer maintains the same structure as the original self-attention layer and is integrated before the cross-attention layer in each transformer block of the UNet, as depicted in Figure~\ref{fig:trans_block}.
In the cross-domain attention layer, the keys and values from the normal and color image domains are combined and processed through attention operations. This design ensures that the generations of color images and normal maps are closely correlated, thus promoting geometric consistency between the two domains.

\subsection{Textured Mesh Extraction}
\label{sec:mesh-extraction}
To extract explicit 3D geometry from 2D normal maps and color images, we optimize a neural implicit signed distance field (SDF) to amalgamate all 2D generated data. Unlike alternative representations like meshes, SDF offers compactness and differentiability, making them ideal for stable optimization.

Nonetheless, adopting existing SDF-based reconstruction methods, such as NeuS~\cite{wang2021neus}, proves unviable. These methods were tailored for real-captured images and necessitate dense input views. In contrast, our generated views are relatively sparse, and the generated normal maps and color images may exhibit subtle inaccurate predictions of some pixels. Regrettably, these errors accumulate during the geometry optimization, leading to distorted geometries, outliers, and incompleteness.
To overcome the challenges above, we propose a novel geometric-aware optimization scheme. 

\input{figures/gso_compare}

\noindent
\textbf{Optimization Objectives.}
With the obtained normal maps $G_{0:N}$ and color images $H_{0:N}$, we first leverage segmentation models to segment the object masks $M_{0:N}$ from the normal maps or color images. 
Specifically, we perform the optimization by randomly sampling a batch of pixels and their corresponding rays in world space $P=\left\{g_k, h_k, m_k, \mathbf{v}_k\right\}$, where $g_k$ is normal value of the $k_{th}$ sampled pixel, $h_k$ is color value of the $k_{th}$ pixel, $m_k \in \{0,1\}$ is mask value of the $k_{th}$ pixel, and $\mathbf{v}_k$ is the direction of the corresponding sampled $k_{th}$ ray, from all views at each iteration.

The overall objective function is defined as 
\begin{equation}
\begin{aligned}
\mathcal{L} &= \mathcal{L}_{normal} + \mathcal{L}_{rgb} + \mathcal{L}_{mask} \\
&+ \mathcal{R}_{eik} + \mathcal{R}_{sparse} + \mathcal{R}_{smooth},
\end{aligned}
\end{equation}
where $\mathcal{L}_{normal}$ denotes the normal loss term that will be discussed later, $\mathcal{L}_{rgb}$ denotes a MSE loss term that calculates the errors between rendered colors $\hat{h}_k$ and generated colors $h_k$, $\mathcal{L}_{mask}$ denotes a binary cross-entropy loss term that calculating errors between the rendered mask $\hat{m}_k$ and the generated mask $m_k$, $\mathcal{R}_{eik}$ denotes eikonal regularization term that encourages the magnitude of the SDF gradients to be unit length, $\mathcal{L}_{sparse}$ denotes a sparsity regularization term that avoid floaters of SDF, and $\mathcal{L}_{smooth}$ denotes a 3D smoothness regularization term that enforces the SDF gradients to be smooth in 3D space.

\noindent
\textbf{Geometry-aware Normal Loss.} Thanks to the differentiable nature of SDF representation, we can easily extract normal values $\hat{g}$ of the optimized SDF via calculating the second-order gradients of SDF. 
We maximize the similarity of the normal of SDF $\hat{g}$ and our generated normal $g$ to provide 3D geometric supervision.
To tolerate trivial inaccuracies of the generated normals from different views, we introduce a geometry-aware normal loss:

\begin{equation}
\begin{split}
    \mathcal{L}_{normal} &= \frac{1}{\sum w_k} \sum w_k \cdot e_{k}\\
e_k &=\left(1-\cos \left(\hat{g}_k, g_k\right)\right), 
\end{split}
\end{equation}
where $e_k$ is error between the normal of SDF $\hat{g}_k$ and the generated normal $g_k$ for the $k_{th}$ sampled ray, $\cos(\cdot, \cdot)$ denotes cosine function, and 
$w_k$ is a geometric-aware weight defined as

\begin{equation}
    w_k= \begin{cases}0, & \cos \left(\mathbf{v}_k, \mathbf{g}_k\right)>\epsilon \\
\exp \left( | \cos \left(\mathbf{v}_k, \mathbf{g}_k\right) | \right), & \cos \left(\mathbf{v}_k, \mathbf{g}_k\right) \leq \epsilon\end{cases}.
\end{equation}
Here $\exp(\cdot)$ denotes exponential function, $|\cdot|$ denotes absolute function, $\epsilon$ is a negative threshold closing to zero, and we measure the cosine value of the angle between the generated normal $g_k$ and the $k_{th}$ ray's viewing direction $\mathbf{v}_k$. 

The design rationale behind this approach lies in the orientation of normals, which are deliberately set to face outward, while the viewing direction is inward-facing. This configuration ensures that the angle between the normal vector and the viewing ray remains not less than $90^{\circ}$. A deviation from this criterion would imply inaccuracies in the generated normals.

Furthermore, it's worth noting that a 3D point on the optimized shape can be visible from multiple distinct viewpoints, thereby being influenced by multiple normals corresponding to these views. However, if these multiple normals do not exhibit perfect consistency, the geometric supervision may become somewhat ambiguous, leading to imprecise geometry.
To address this issue, rather than treating normals from different views equally, we introduce a weighting mechanism. We assign higher weights to normals that form larger angles with the viewing rays. This prioritization enhances the accuracy of our geometric supervision process.

\input{figures/condx}

\noindent
\textbf{Outlier-dropping Losses.}
Besides the normal loss, mask loss and color loss are also adopted for optimizing geometry and appearance.
However, it is inevitable that there exist some inaccuracies in the masks and color images, which will accumulate in the optimization and thus cause noisy surfaces and holes.

To mitigate the issues, we employ a simple yet effective strategy named outlier-dropping loss.
Taking the color loss calculation as an example, instead of simply summing up the color errors of all sampled rays at each iteration, we first sort these errors in a descending order and then discard the top largest errors according to a predefined percentage.
This approach is motivated by the fact that erroneous predictions lack sufficient consistency with other views, making them less amenable to effective minimization during optimization, and they often result in large errors. By implementing this strategy, the optimized geometry can eliminate incorrect isolated geometries and distorted textures.

%% file: figures/color-compare.tex
\begin{figure*}[htp!]
\centering
\includegraphics[width=\linewidth]{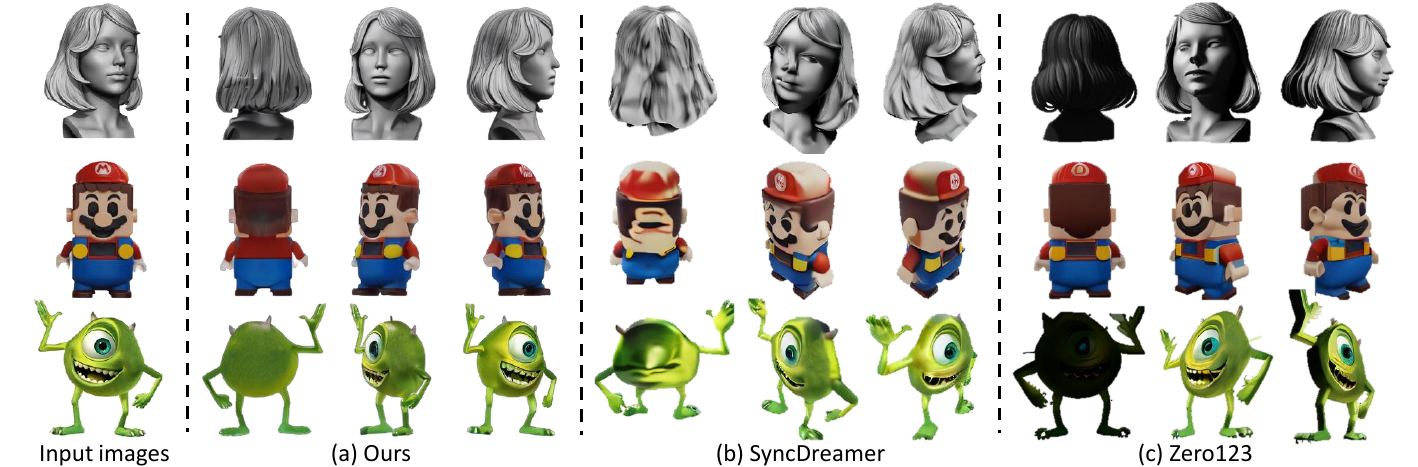}
\caption{The qualitative comparisons with baseline models on synthesized multi-view color images.}
\label{color-compare}	
\end{figure*}

%% file: figures/trans-block.tex
\begin{figure}[tp!]
\centering
\includegraphics[width=0.8\linewidth]{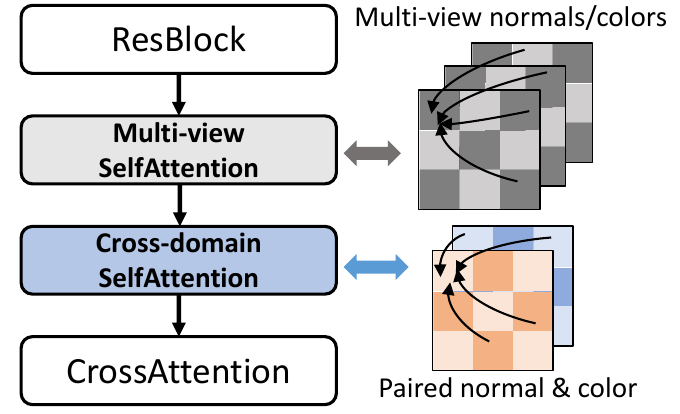}
\caption{The illustration of the structure of the multi-view cross-domain transformer block.}
\label{fig:trans_block}	
\end{figure}

%% file: figures/cartoon.tex
\begin{figure*}[t]
\centering
\includegraphics[width=\linewidth]{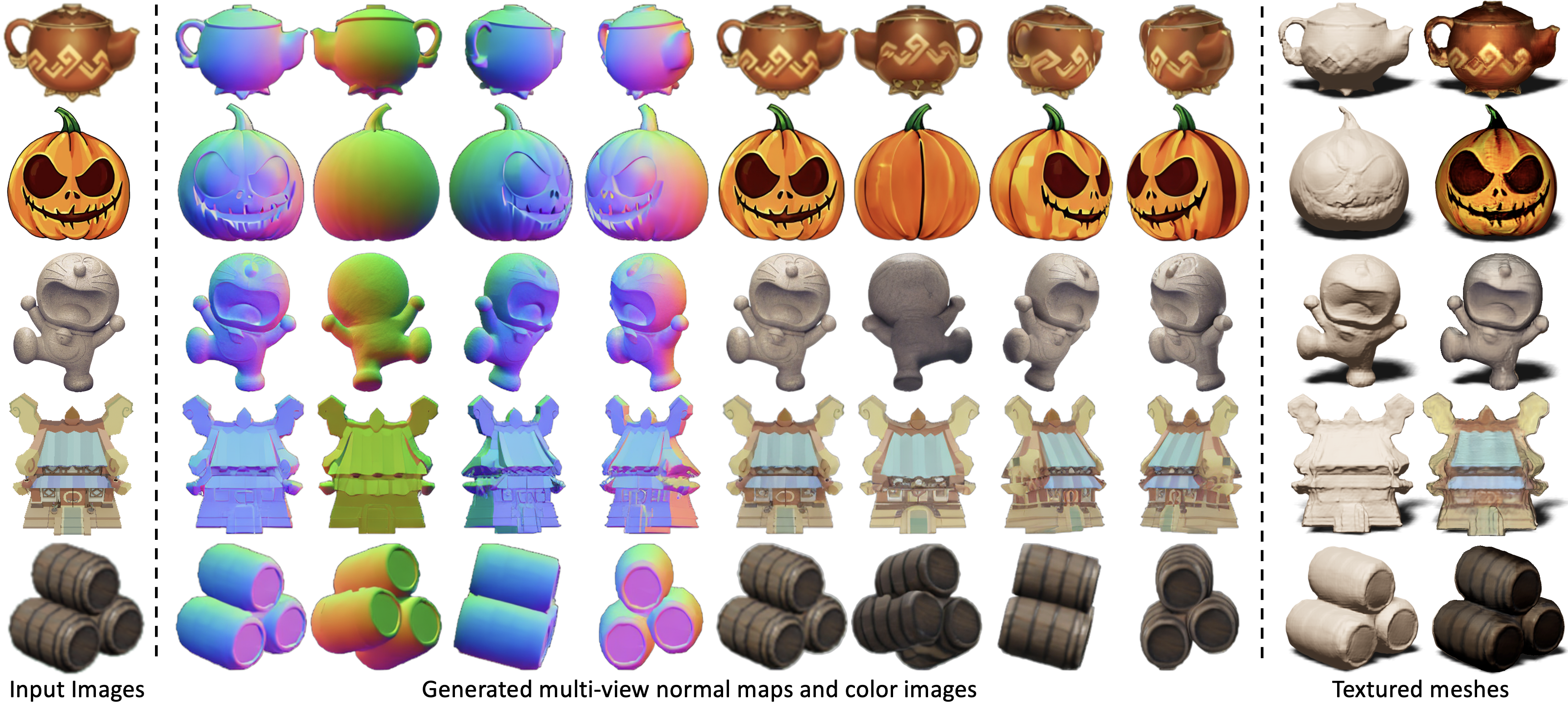}
\caption{The qualitative results of \textit{Wonder3D} on various styles of images.}
\label{fig:cartoon}	
\end{figure*}

%% file: figures/gso_compare.tex
\begin{figure*}[t]
\centering
\includegraphics[width=\linewidth]{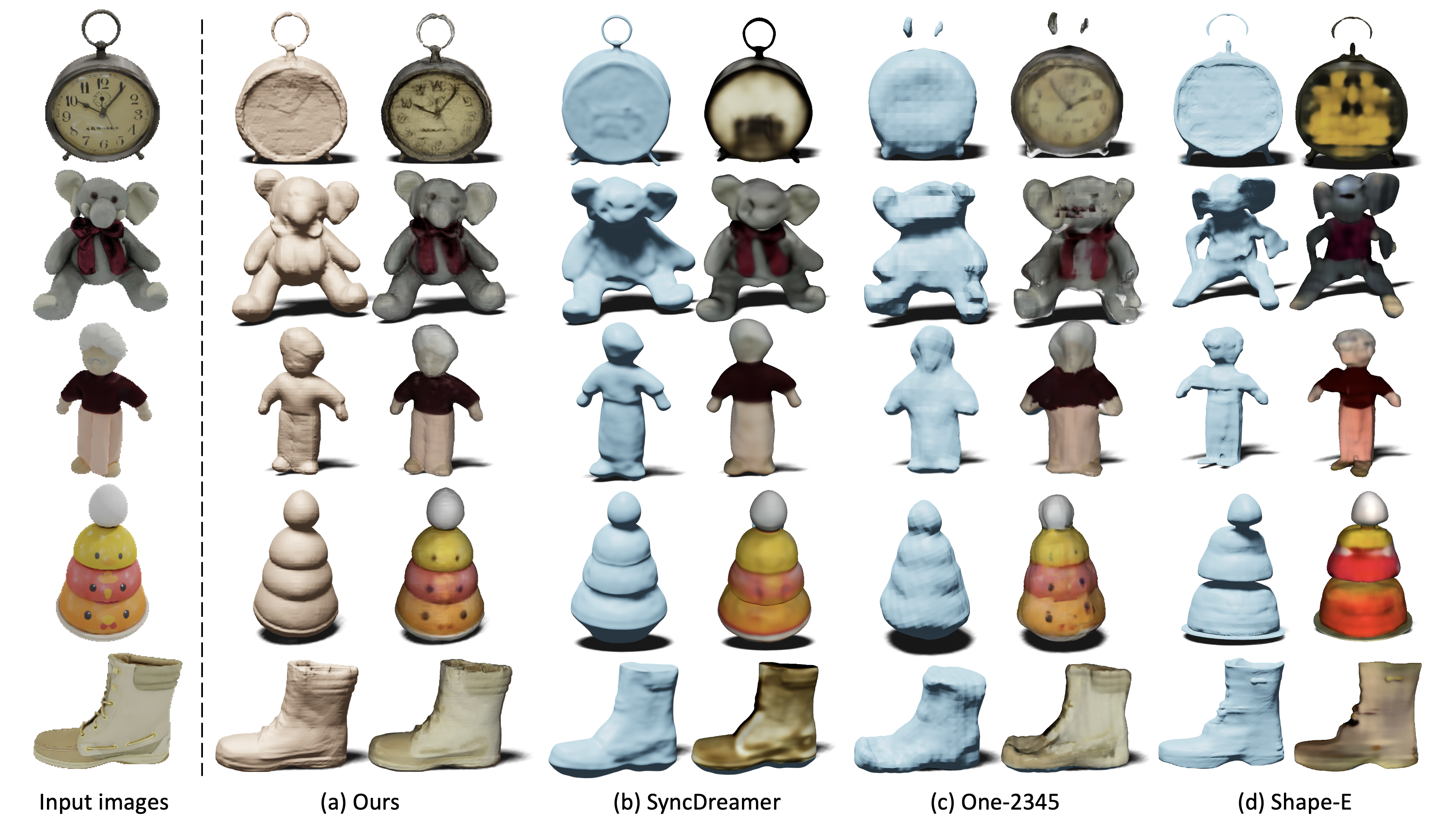}
\caption{The qualitative comparisons with baseline methods on GSO~\cite{downs2022google} dataset in terms of the reconstructed textured meshes.}
\label{gso_compare}	
\end{figure*}

%% file: figures/condx.tex
\begin{figure*}[tp!]
\centering
\includegraphics[width=\linewidth]{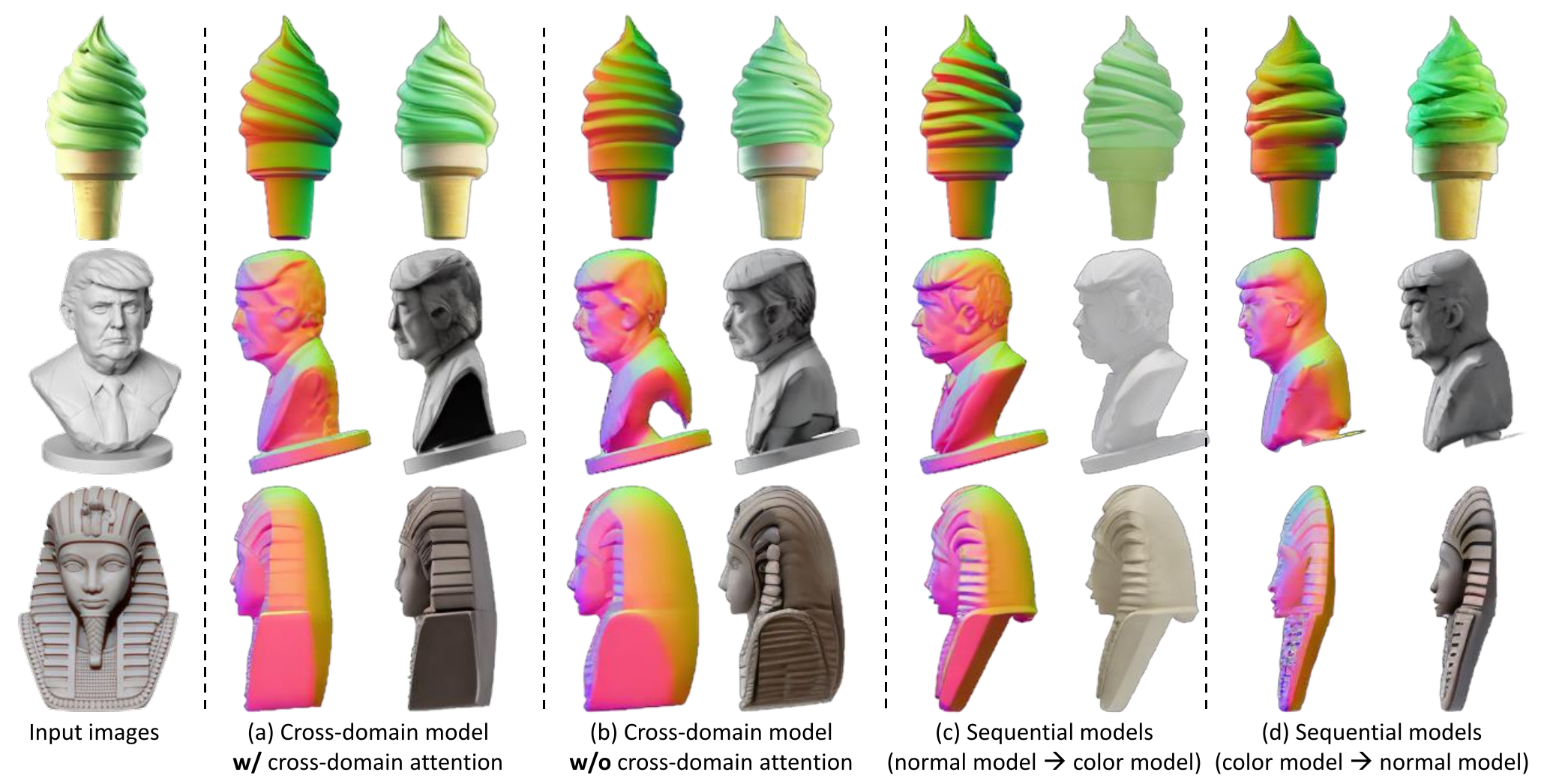}
\caption{Ablation studies on different cross-domain diffusion schemes. 
}

\label{fig:ablation_condx}	
\end{figure*}

%% file: sections/5_exp.tex
\section{Experiments}
\subsection{Implementation Details}
We train our model on the LVIS subset of the Objaverse dataset~\cite{deitke2023objaverse}, which comprises approximately 30,000+ objects following a cleanup process. Surprisingly, even with fine-tuning on this relatively small-scale dataset, our method demonstrates robust generalization capabilities.
To create the rendered multi-view dataset, we first normalized each object to be centered and of unit scale. Then we render normal maps and color images from six views, including the front, back, left, right, front-right, and front-left views, using Blenderproc~\cite{Denninger2023}. Additionally, to enhance dataset diversity, we applied random rotations to the 3D assets during the rendering process.

We fine-tune our model starting from the Stable Diffusion Image Variations Model, which has previously been fine-tuned with image conditions. 
We retain the optimizer settings and $\epsilon$-prediction strategy from the previous fine-tuning.
During fine-tuning, we use a reduced image size of 256 $\times$ 256 and a total batch size of 512 for training. The fine-tuning process involves training the model for 30,000 steps. This entire training procedure typically requires approximately 3 days on a cluster of 8 Nvidia Tesla A800 GPUs. To reconstruct 3D geometry from the 2D representations, our method is built on the instant-NGP based SDF reconstruction method~\cite{instant-nsr-pl}.

\input{figures/ablation-normalfusion}

\subsection{Baselines}
We adopt Zero123~\cite{liu2023zero}, RealFusion~\cite{melas2023realfusion}, Magic123~\cite{qian2023magic123}, One-2-3-45~\cite{liu2023one}, Point-E~\cite{nichol2022point}, Shap-E~\cite{jun2023shap} and a recent work SyncDreamer~\cite{liu2023syncdreamer} as baseline methods. %
Given an input image, zero123 is capable of generating novel views of arbitrary viewpoints, and it can be incorporated with SDS loss~\cite{poole2022dreamfusion} for 3D reconstruction (we adopt the implementation of ThreeStudio~\cite{threestudio2023}).
RealFusion~\cite{melas2023realfusion} and Magic123~\cite{qian2023magic123} leverage Stable Diffusion~\cite{rombach2022high} and SDS loss for single-view reconstruction.
One-2-3-45~\cite{liu2023one} directly predict SDFs via SparseNeuS~\cite{long2022sparseneus} by taking 
the generated multiple images of Zero123~\cite{liu2023zero}.
Point-E~\cite{nichol2022point} and Shap-E~\cite{jun2023shap} are 3D generative models trained on a large internal OpenAI 3D dataset, both of which are able to convert a single-view image into a point cloud or an implicit representation. SyncDreamer\cite{liu2023syncdreamer} aims to generate multi-view consistent images from a single image for deriving 3D geometry.

\subsection{Evaluation Protocol}
\noindent
\textbf{Evaluation Datasets.} Following prior research~\cite{liu2023zero,liu2023syncdreamer}, we adopt the Google Scanned Object dataset~\cite{downs2022google} for our evaluation, which includes a wide variety of common everyday objects. 
Our evaluation dataset matches that of SyncDreamer~\cite{liu2023syncdreamer}, comprising 30 objects that span from everyday items to animals.
For each object in the evaluation set, we render an image with a size of 256×256, which serves as the input. Additionally, to assess the generalization ability of our model, we include some images with diverse styles collected from the internet in our evaluation.

\noindent
\textbf{Metrics.} To evaluate the quality of the single-view reconstructions, we adopt two commonly used metrics Chamfer Distances (CD) and Volume IoU between ground-truth shapes and reconstructed shapes.
Since different methods adopt various canonical systems, we first align the generated shapes to the ground-truth shapes before calculating the two metrics.
Moreover, we adopt the metrics PSNR, SSIM~\cite{wang2004image} and LPIPS~\cite{zhang2018unreasonable} for evaluating the generated color images.

\subsection{Single View Reconstruction}

\input{tables/reconstruction}
We evaluate the quality of the reconstructed geometry of different methods. The quantitative results are summarized in Table~\ref{tab:recon}, and the qualitative comparisons are presented in Fig.~\ref{gso_compare}.
Shap-E~\cite{jun2023shap} tends to produce incomplete and distorted meshes. SyncDreamer~\cite{liu2023syncdreamer} generates shapes that are roughly aligned with the input image but lack detailed geometries, and the texture quality is subpar.
One-2-3-45~\cite{liu2023one} attempts to reconstruct meshes from the multiview-inconsistent outputs of Zero123~\cite{liu2023zero}. While it can capture coarse geometries, it loses important details in the process.
In comparison, our method stands out by achieving the highest reconstruction quality, both in terms of geometry and textures.

\subsection{Novel View Synthesis}

\input{tables/nvs}
We evaluate the quality of novel view synthesis for different methods. The quantitative results are presented in Table~\ref{tab:nvs}, and the qualitative results can be found in Figure~\ref{color-compare}. 
Zero123~\cite{liu2023zero} produces visually reasonable images, but they lack multi-view consistency since it operates on each view independently. Although SyncDreamer~\cite{liu2023zero} introduces a volume attention scheme to enhance the consistency of multi-view images, their model is sensitive to the elevation degrees of the input images and tends to produce unreasonable results.
In contrast, our method is capable of generating images that not only exhibit semantic consistency with the input image but also maintain a high degree of consistency across multiple views in terms of both colors and geometry.

\subsection{Discussions}
\label{sec:discussion}
In this section, we conduct a set of studies to verify the effectiveness of our designs as well as the properties of the method. 

\noindent
\textbf{Cross-Domain Diffusion.}
\input{figures/ablation-mvattn}
To validate the effectiveness of our proposed cross-domain diffusion scheme, we study the following settings: (a) cross-domain model with cross-domain attention; 
(b) cross-domain model without cross-domain attention; 
(c) sequential model rgb-to-normal: first train a multi-view color diffusion model then train a multi-view normal diffusion model conditioned on the previously generated color images; (d) sequential model normal-to-rgb: first train a multi-view normal diffusion model then train a multi-view color diffusion model conditioned on the previously generated normal images.

As shown in (a) and (b) of Figure~\ref{fig:ablation_condx}, it's evident that the cross-domain attentions significantly enhance the consistency between color images and normals, particularly in terms of the detailed geometries of objects like the ice-cream and Pharaoh sculpture. 
From (c) and (d) of Figure~\ref{fig:ablation_condx}, while the normals and color images generated by sequential models maintain some consistency, their results suffer from performance drops. 

For the sequential model rgb-to-normal, conditioning on the separately generated normal maps, the generated color images exhibit color aberrations in comparison to the input image, as shown in (c) of Figure~\ref{fig:ablation_condx}. 
Conversely, for the sequential model normal-to-rgb, conditioning on the separately generated color images, the normal maps give unreasonable geometry, as illustrated in (d) of Figure~\ref{fig:ablation_condx}. These experiments demonstrate that jointly predicting normal maps and color images through the cross-domain attention mechanism can facilitate a comprehensive perception of information from different domains. We also speculate that in the context of sequential models, the generated color images or normal maps of stage 1 may exhibit a minor domain gap to the ground truth data trained in stage 2. Therefore, compared to sequential prediction, the cross-domain approach proves to be more effective in enhancing the quality of each domain as well as the overall prediction.

\noindent
\textbf{Multi-view Consistency.}
We conducted an analysis of the effectiveness of the multi-view attention mechanism, as illustrated in Figure~\ref{fig:ablation_mvattn}. Our findings show that the multi-view attention greatly enhances the 3D consistency of the generated multi-view images, particularly for the rear views. In the absence of the multi-view attention, the color images of the rear views exhibited unrealistic predictions.

\noindent
\textbf{Normal Fusion.}
To assess the efficacy of our normal fusion algorithm, we conducted experiments using the complex lion model, which is rich in geometric details, as illustrated in Figure \ref{fig:normalfusion}. The baseline model's surfaces exhibited numerous holes and noises. Utilizing either the geometry-aware normal loss or the outlier-dropping loss helps mitigate the noisy surfaces. Finally, combining both strategies yields the best performance, resulting in clean surfaces while preserving detailed geometries.

\noindent
\textbf{Generalization.}
To demonstrate the generalization capability of our method, we conducted evaluations using diverse image styles, including sketches, cartoons, and images of animals, as shown in Figure~\ref{fig:cartoon} and Figure~\ref{fig:animals}. Despite variations in lighting effects and geometric complexities among these images, our method consistently generated multi-view normal maps and color images, ultimately yielding high-quality geometries.

\input{figures/animals}

%% file: figures/ablation-normalfusion.tex
\begin{figure*}[t]
\centering
\includegraphics[width=\linewidth]{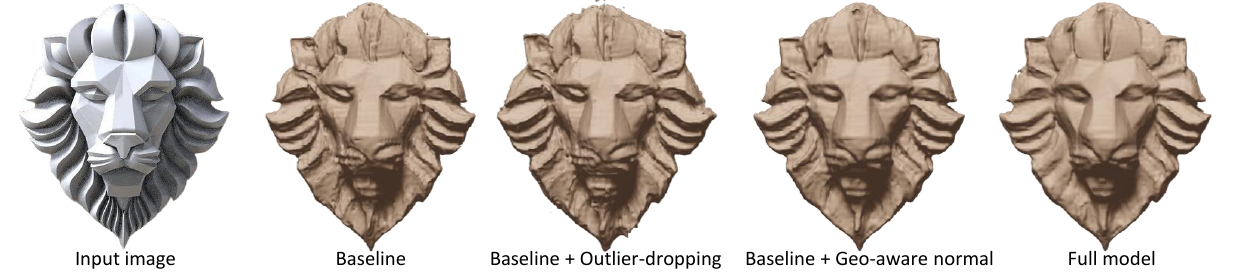}
\caption{Ablation study on the strategies in the mesh extraction module: geometry-aware normal loss and outlier-dropping strategy.
}
\label{fig:normalfusion}	
\end{figure*}

%% file: tables/reconstruction.tex
\begin{table}[]
    \centering
    \resizebox{0.85\linewidth}{!}{
    \begin{tabular}{lcc}
       \toprule
       Method  & Chamfer Dist.$\downarrow$ & Volume IoU$\uparrow$ \\
       \midrule
       Realfusion~\cite{melas2023realfusion}    
       & 0.0819  & 0.2741   \\
       Magic123~\cite{qian2023magic123}
       & 0.0516 &  0.4528 \\
       One-2-3-45~\cite{liu2023one}    
       & 0.0629 &  0.4086 \\
       Point-E~\cite{nichol2022point}    
       & 0.0426 & 0.2875 \\
       Shap-E~\cite{jun2023shap}    
       & 0.0436 &  0.3584  \\
       Zero123~\cite{liu2023zero}    
       & 0.0339 &  0.5035 \\
       SyncDreamer~\cite{liu2023syncdreamer}   
       &  {0.0261}  &  {0.5421}   \\
        Ours    
       &  \textbf{0.0199}  &  \textbf{0.6244}   \\
       \bottomrule
    \end{tabular}
    }
    \caption{Quantitative comparison with baseline methods. We report Chamfer Distance and Volume IoU on the GSO~\cite{downs2022google} dataset.}
    \label{tab:recon}
\end{table}

%% file: tables/nvs.tex
\begin{table}[]
    \centering
    \resizebox{0.75\linewidth}{!}{
    \begin{tabular}{lccc}
       \toprule
       Method  & PSNR$\uparrow$ & SSIM$\uparrow$ & LPIPS$\downarrow$  \\
       \midrule
    Realfusion~\cite{melas2023realfusion}    
       & 15.26 & 0.722 & 0.283   \\
       Zero123~\cite{liu2023zero}    
       & 18.93 & 0.779 & 0.166   \\
       SyncDreamer~\cite{liu2023syncdreamer}   
       & {20.05} & {0.798} & {0.146} \\
       Ours    
       & \textbf{26.07} & \textbf{0.924} & \textbf{0.065} \\
       \bottomrule
    \end{tabular}
    }
    \caption{The quantitative comparison in novel view synthesis. We report PSNR, SSIM~\cite{wang2004image}, LPIPS~\cite{zhang2018unreasonable} on the GSO~\cite{downs2022google} dataset.}
    \label{tab:nvs}
\end{table}

%% file: figures/ablation-mvattn.tex
\begin{figure}[tp!]
\centering
\includegraphics[width=\linewidth]{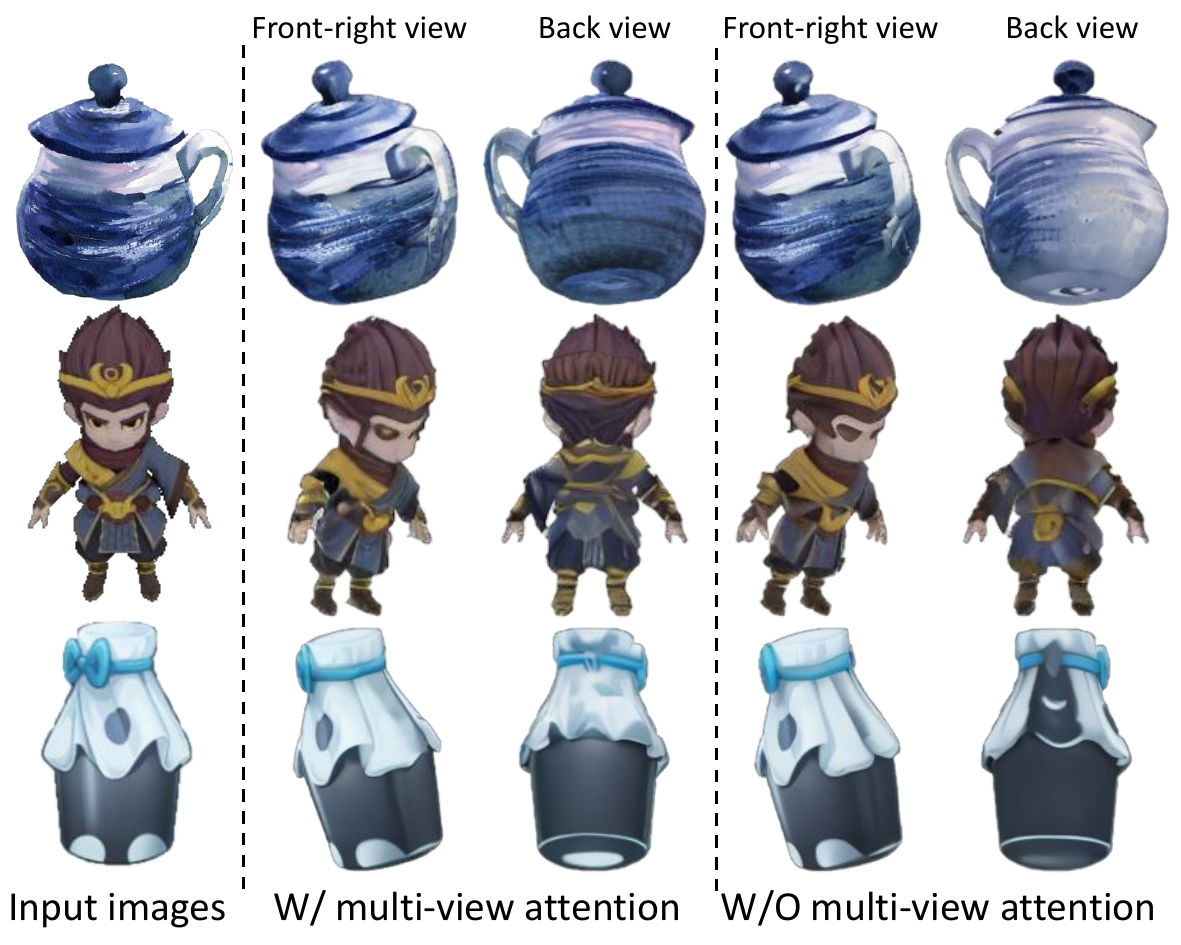}
\caption{Ablation study on multi-view attention.
}
\label{fig:ablation_mvattn}	
\end{figure}

%% file: figures/animals.tex
\begin{figure}[tp!]
\centering
\includegraphics[width=\linewidth]{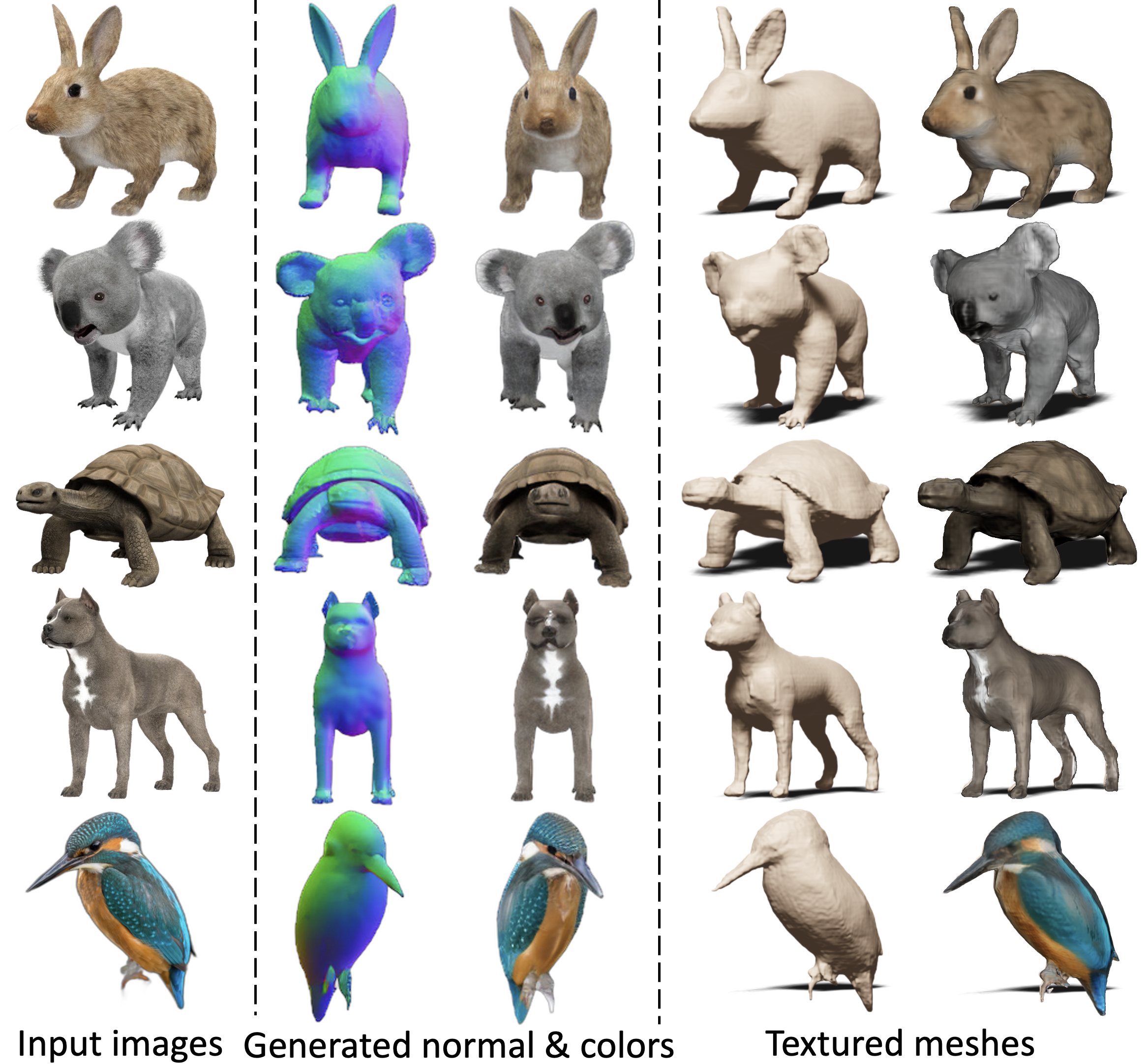}
\caption{The qualitative results of \textit{Wonder3D} on various animal objects. 
}
\label{fig:animals}	
\end{figure}

%% file: sections/6_conclusion.tex
\section{Conclusions and Future Works}

\noindent
\textbf{Conclusions.} In this paper, we present \textit{Wonder3D}, an innovative approach designed for efficiently generating high-fidelity textured meshes from single-view images. When provided with a single image, \textit{Wonder3D} initiates the process by generating consistent multi-view normal maps and paired color images. Subsequently, it utilizes a novel normal fusion algorithm to extract highly-detailed geometries from these multi-view 2D representations. Experimental results demonstrate that our method upholds good efficiency and robust generalization, and delivers high-quality geometry.

\noindent
\textbf{Future Works.}
While \textit{Wonder3D} has demonstrated promising performance in reconstructing 3D geometry from single-view images, there are still some limitations that the current framework does not fully address. First, the current implementation of \textit{Wonder3D} only produces normals and color images from six views. This limited number of views makes it challenging for our method to accurately reconstruct objects with very thin structures and severe occlusions.
Additionally, expanding \textit{Wonder3D} to incorporate more views would demand increased computational resources during training. To address this issue, \textit{Wonder3D} may benefit from leveraging more efficient multi-view attention mechanisms to handle a greater number of views effectively.

\section*{Acknowledgements}
Thanks for the GPU support from VAST, the valuable suggestions from Wei Yin, the help in data rendering from Dehu Wang.